\documentclass[letterpaper, 10 pt, conference]{ieeeconf}
\IEEEoverridecommandlockouts
\usepackage[space, compress, sort]{cite}
\usepackage{amsmath, amsfonts}

\usepackage{amsthm, amssymb}
\usepackage[ruled,vlined]{algorithm2e}
\usepackage{mathtools}
\usepackage{tabularx}
\usepackage{graphicx}
\usepackage{enumerate}
\usepackage{float}
\usepackage{url}
\usepackage{verbatim}
\usepackage{booktabs} 
\usepackage{multirow}
\usepackage[linkcolor=black,citecolor=black,urlcolor=black,colorlinks=true]{hyperref}
\usepackage{bm}
\usepackage{algpseudocode}
\usepackage{xcolor}
\usepackage{tcolorbox}
\usepackage{url} 
\usepackage{hyperref}
\usepackage{tabularx}
\usepackage{float}
\usepackage{makecell}
                                    \title{\LARGE \bf
SPAR: Scalable LLM-based PDDL Domain Generation \\
for Aerial Robotics}
\author{Songhao Huang$^{*1}$, Yuwei Wu$^{*1}$, Guangyao Shi$^{2}$, Gaurav S. Sukhatme$^{2}$, Vijay Kumar$^{1}$
\thanks{$^{*}$Equal contribution. $^{1}$ Songhao Huang, Yuwei Wu, and Vijay Kumar are with the GRASP Lab, University of Pennsylvania, Philadelphia, PA 19104, USA. Email: \texttt{\{songhaoh, yuweiwu, kumar\}@seas.upenn.edu}.}%
\thanks{$^{2}$ Guangyao Shi and Gaurav S. Sukhatme are with the Department of Computer Science, University of Southern California, Los Angeles, CA 90089, USA. Email: \texttt{\{shig, gaurav\}@usc.edu}.}
\thanks{We also thank Jonathan Diller for providing helpful comments.}
}

\begin{document}
\maketitle
\begin{abstract}
We investigate the problem of automatic domain generation for the Planning Domain Definition Language (PDDL) using Large Language Models (LLMs), with a particular focus on unmanned aerial vehicle (UAV) tasks. 
Although PDDL is a widely adopted standard in robotic planning, manually designing domains for diverse applications such as surveillance, delivery, and inspection is labor-intensive and error-prone, which hinders adoption and real-world deployment.
To address these challenges, we propose {\sc Spar}, a framework that leverages the generative capabilities of LLMs to automatically produce valid, diverse, and semantically accurate PDDL domains from natural language input.
To this end, we first introduce a systematically formulated and validated UAV planning dataset, consisting of ground-truth PDDL domains and associated problems, each paired with detailed domain and action descriptions. 
Building on this dataset, we design a prompting framework that generates high-quality PDDL domains from language input. 
The generated domains are evaluated through syntax validation, executability, feasibility, and interpretability. 
Overall, this work demonstrates that LLMs can substantially accelerate the creation of complex planning domains, providing a reproducible dataset and evaluation pipeline that enables application experts without prior experience to leverage it for practical tasks and advance future research in aerial robotics and automated planning.

\end{abstract}

\IEEEpeerreviewmaketitle

\section{Introduction}

The Planning Domain Definition Language (PDDL)~\cite{aeronautiques1998pddl} is a fundamental formalism in automated planning, widely used in diverse areas such as robot task planning, logistics, and human-robot interactions~\cite{ramirez2018integrated, han2024interpret, Garrett_Lozano-Pérez_Kaelbling_2020}. 
Despite its expressive power, creating correct PDDL domain and problem files is often time-consuming and requires significant expertise, even for experienced users, which limits its broader adoption in planning systems for real-world applications, particularly in dynamic and safety-critical domains like Unmanned Aerial Vehicle (UAV) operations. 
These domains involve complex three-dimensional environments, GPS-denied conditions, external disturbances, and limited observations, making specification technically non-trivial.

Recent advances in Large Language Models (LLMs) offer promising opportunities to automate the generation of PDDL from natural language descriptions. 
Prior work has demonstrated notable success in generating PDDL problem files~\cite{liu2023llmp, dagan2023dynamicplanningllm}, which specify concrete initial states and goals. 
However, generating domain files remains substantially more challenging. 
Unlike problem files, domain files define reusable action schemas that model the structure, dynamics, and constraints of a planning environment, requiring abstraction, consistency, and logical reasoning. 
Consequently, automating domain generation and transfer has attracted increasing research attention~\cite{Strobel2020, smirnov2024generatingconsistentpddldomains, khandelwal2024pddlfusetoolgeneratingdiverse, 11072193, oswald2024largelanguagemodelsplanning}. 
Nevertheless, these reasoning capabilities remain difficult for LLMs to reliably demonstrate.

\begin{figure}[t]
    \centering
    \includegraphics[width=0.98\columnwidth]{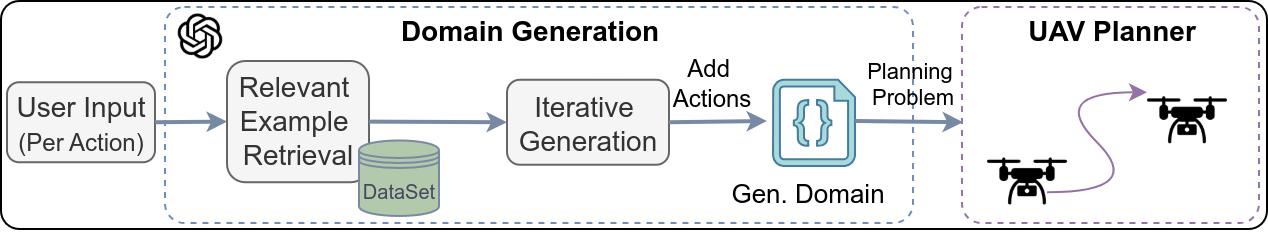}
    \vspace{-1em}
    \caption{The {\sc Spar} framework for domain generation and planning.}
    \label{fig:overview}
    \vspace{-2em}
\end{figure}

Moreover, the success of LLM-based domain generation is highly sensitive to the amount and quality of information provided.
Existing studies often assume access to detailed task descriptions, expert knowledge, or ground-truth domain models\cite{mahdavi2024leveraging, zhang-etal-2024-proc2pddl}.
While such conditions yield promising results, they do not reflect realistic scenarios where non-experts provide ambiguous or incomplete descriptions of unfamiliar tasks, or they don't have access to real-world environments.
In these cases, LLM-generated domain files frequently exhibit syntactic and semantic errors, logical inconsistencies, and incomplete action definitions. 
This discrepancy raises a critical question: \textbf{How can we empower LLMs to automatically construct reliable PDDL domain definitions for novel tasks, even when faced with unfamiliar actions and limited prior knowledge?}

To address these challenges, we study UAV planning as a representative and demanding domain involving complex tasks with diverse constraints. To enable a systematic evaluation of LLMs in generating diverse and previously unseen planning domains, our first step involves the following procedures: (1) define a domain complexity metric that quantifies difficulty across planning domains; (2) create a new PDDL domain evaluation dataset, which covers hand-crafted and LLM-generated domains from diverse sources to ensure novelty relative to LLM training data. This dataset includes a wide range of UAV scenarios (e.g., navigation, inspection, delivery) and diverse actions (e.g., detection, loading, recharging); and (3) defines a set of evaluation metrics for assessing both syntactic and semantic correctness.

Building on this foundation, we introduce {\sc Spar} (Fig.~\ref{fig:overview}), a generation framework designed to address two common issues in LLM-generated domains: syntax errors and semantic errors. 
To mitigate syntax errors, the framework adopts an action-by-action generation process with a functional syntax checker in a feedback loop for iterative correction. To mitigate semantic errors, it incorporates chain-of-thought (CoT) reasoning to guide step-by-step action generation. Furthermore, inspired by retrieval-augmented generation (RAG)~\cite{10.5555/3495724.3496517}, the framework retrieves the most relevant examples from a curated dataset to support in-context reasoning.  
Our main contributions are as follows:
\begin{itemize}
    \item We present a \textit{first} multi-domain PDDL dataset for UAV applications, consisting of 30 domain-problem pairs with aligned natural language descriptions. The dataset includes both logical and numeric domains designed to be unseen for current LLMs.
    \item We propose a domain generation framework that combines CoT prompting, relevant-example retrieval, and an action-by-action feedback generation schema to improve generation ability. 
    \item We benchmark multiple LLMs and prompting strategies on our dataset across multiple evaluation metrics and validate the framework in  UAV simulation experiments, demonstrating the practical feasibility for downstream planning and execution.

\end{itemize}

\section{Related Work}

\subsection{Using LLMs in Task Planning}

LLMs have recently been explored in various capacities to assist with task planning in symbolic domains.

\paragraph{LLMs as task planners} 

One line of research attempts to use LLMs as task planners to generate valid, executable plans directly from natural language instructions. However, naively prompting LLMs often leads to issues with correctness and completeness, especially for complex or long-horizon tasks~\cite{yao2022react, aghzal2025surveylargelanguagemodels, cao2025largelanguagemodelsplanning, choi2024lotabench}.
To mitigate these issues, recent works propose augmenting LLMs with high-level reasoning, optimization methods, or domain-specific priors to improve reliability~\cite{10.1609/aaai.v38i18.30006, 2024hierarchicalllms}. In the context of PDDL-based planning, few-shot prompting has been explored~\cite{10.1609/aaai.v38i18.30006}, showing that careful prompt engineering and example selection can substantially improve the quality and validity of generated plans.  

\paragraph{LLMs for problem or goal generation}  

Since LLMs struggle to serve directly as planners with correctness guarantees, an alternative approach is to use them for generating PDDL problem specifications from natural language, given a known domain model.
These generated problems can then be solved using traditional structured and symbolic planning approaches.
For example, LLM+P~\cite{liu2023llmp} is designed to generate PDDL problem instances in robotic settings by conditioning on both the domain file and natural language input.
Similarly, Xie et al.~\cite{xie2023translatingnaturallanguageplanning} showed that LLMs are capable of converting high-level natural language goals into structured, machine-interpretable representations. 
These efforts illustrate how LLMs can serve as problem generators, bridging flexible language interfaces with robust symbolic planning systems.

However, all of the above methods assume the availability of a predefined symbolic world model, specifically a PDDL domain. 
Therefore, \textbf{PDDL domain generation} becomes a more foundational challenge, since the accuracy and utility of problem generation and plan synthesis are determined by the quality of the underlying domain model.

\subsection{Planning Domain Generation}

Generating planning domains with LLMs requires strong reasoning capabilities. 
Prior work has explored several strategies to improve the quality of LLM-generated PDDL. 
Guan et al.~\cite{guan2023leveraging} employed in-context learning with curated examples to improve syntactic correctness, while Zhang et al.~\cite{zhang-etal-2024-proc2pddl} introduced Zone of Proximal Development (ZPD) prompting to enhance reasoning and support open-domain generation. 
Mahdavi et al.~\cite{mahdavi2024leveraging} further proposed an environment-feedback loop to refine domains based on semantic similarity.  

A key challenge in domain generation lies in the quantity and quality of input information. 
Providing additional knowledge, such as predicates~\cite{zhang-etal-2024-proc2pddl,oswald2024largelanguagemodelsplanning} or structural descriptions~\cite{mahdavi2024leveraging} often improves accuracy, but such inputs are not always available in real-world scenarios~\cite{guan2023leveraging}. 
To address this, we leverage RAG to retrieve the most relevant examples for in-context learning under limited input conditions to improve semantic fidelity.  
Feedback mechanisms are also crucial for domain quality. Guan et al.~\cite{guan2023leveraging} proposed a syntax validator with human-in-the-loop refinement, while Smirnov et al.~\cite{smirnov2024generating} and Mahdavi et al.~\cite{mahdavi2024leveraging} automated feedback through consistency checks or environment evaluation. However, these methods often assume access to the ground-truth (GT) environment. Moreover, many validators (e.g., Tarski~\cite{tarski:github:18}) lack support for numeric conditions. Since our dataset includes numeric operators, we extend the validator from Guan et al.~\cite{guan2023leveraging} and incorporate LLM-friendly feedback to improve syntax correctness during action-by-action generation. 

\subsection{UAV Task Planning Benchmark} 
Existing UAV task planning benchmarks primarily focus on a single task type or the UAV platform itself, such as inspection with heterogeneous UAVs\cite{cao2025cooperative}, object detection\cite{ye2025more}, and goal navigation\cite{xiao2025uav}. 
With VLA/VLN, more UAV benchmarks have been proposed in both simulation and real-world settings~\cite{xie2024travelplanner, wang2025uav}. 
To broaden UAV applications, our dataset encompasses not only standard tasks such as navigation, detection, coordination, and exploration, 
but also non-standard scenarios, including aerial transportation with aerial manipulators, flying cars, and others. 
In addition, few works have introduced PDDL-based benchmarks for UAV task planning. 
To bridge this gap, we formulate all task planning problems in PDDL, extending the standard STRIPS formalism~\cite{FIKES1971189} to incorporate numerical cases, thereby making our dataset more representative of real-world conditions.

\begin{figure*}[t]
    \centering
    \vspace{0.5em}
\includegraphics[width=0.98\textwidth]{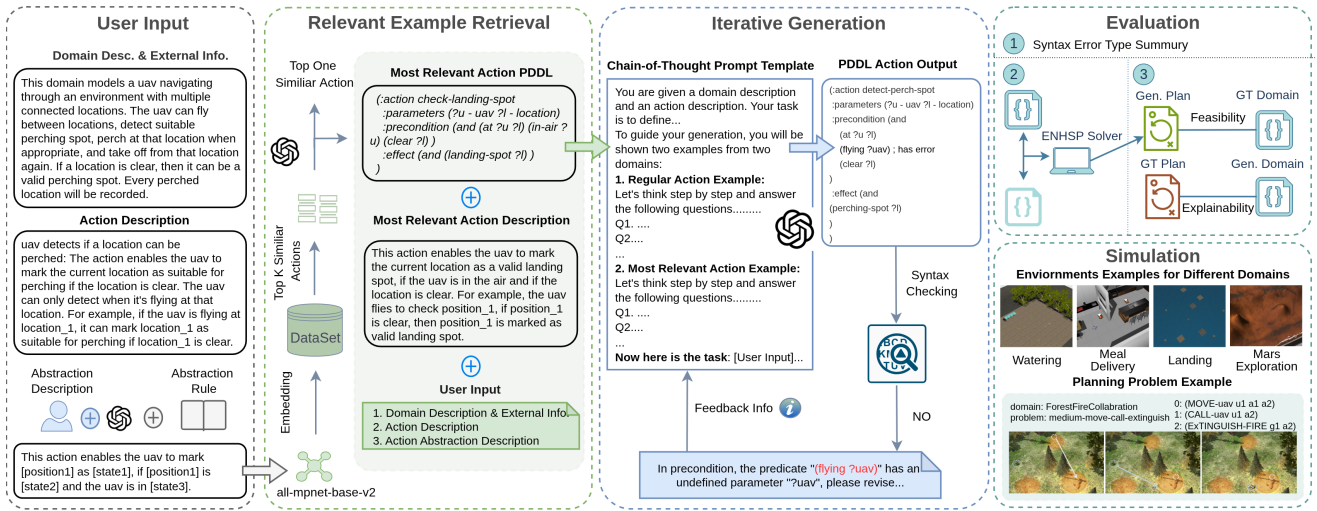}
 \vspace{-0.5em}
    \caption{The {\sc Spar} Domain Generation and Evaluation Pipeline.}
    \label{fig:pipeline}
    \vspace{-1em}
\end{figure*}

\section{Problem Formulation}
\subsection{Preliminaries}
The {PDDL} is a standardized language used in artificial intelligence and robotics to formally describe planning problems. It separates a problem into two parts: the \emph{domain}, which defines the general rules of the world (such as types of objects, predicates, and actions), and the \emph{problem}, which specifies a particular scenario with initial conditions and goals. For example, in a {UAV surveillance task}, the domain might define actions such as \texttt{fly}, \texttt{take-photo}, and \texttt{return-to-base}, along with predicates like \texttt{(at ?uav ?location)} and \texttt{(photo-taken ?location)}. Then, a specific problem could describe a UAV starting at the base, visiting two waypoints, and taking photos at each. Using PDDL, a planner can automatically generate a sequence of actions to fulfill the task, such as \texttt{(fly base waypoint1) $\rightarrow$ (take-photo waypoint1) $\rightarrow$ (fly waypoint1 waypoint2) $\rightarrow$ (take-photo waypoint2) $\rightarrow$ (fly waypoint2 base)}. Such a structured representation makes it easier to design, share, and solve planning tasks across different domains.

\subsection{Problem Statement}
We formally introduce the notation used throughout this paper.
A planning problem \( T \) is defined in PDDL as a tuple \( T = \{D, P\} \), where \( D \) denotes the PDDL domain, which specifies the general planning environment, and \( P \) denotes the PDDL problem, which grounds a specific planning instance within that environment.  
We assume that the problem \( P \) will be available after the domain \( D \) is generated.

The domain \( D \) consists of a set of \textit{fluents} \( F \) and a set of \textit{actions} \( A \). 
Fluents represent the state variables of the environment, typically including both predicates and numeric functions.  
Each action \( a \in A \) is defined by a set of parameters, along with its preconditions \( \text{Pre}(a) \), which must hold before the action is executed, and its effects \( \text{Eff}(a) \), which describe how the environment changes after execution.

We denote the natural language description of the domain as \( N_d \), and the natural language description of the \( i^{th}\) action \( a_i \in A \) by \( N_{a_i} \). 
We say that \( N_{a_i} \) is \emph{incomplete} if it does not explicitly describe all the conditions in \( \text{Pre}(a_i) \) and \( \text{Eff}(a_i) \) using natural language one-by-one.  
Additionally, we denote the declared parameter types and requirements of the domain as \( \text{Extern} \).
Before generation, LLM does not have access to the GT  domain; that is, it has no prior knowledge of the environment's state or actions.

Given a tuple $(N_d, N_A, \text{Extern})$, where some action descriptions $N_{a_i} \in N_A$ may be incomplete, we aim to use LLMs to automatically generate a PDDL domain $\hat{D} = (\hat{F}, \hat{A})$. With a corresponding problem instance $\hat{P}$ (manually specified or LLM-generated), a classical planner $C$ solves the task $\hat{T} = \{\hat{D}, \hat{P}\}$, producing a plan $\textit{Plan} \subseteq \hat{A}$ that transitions the system from the initial state to the goal state.

\section{Approach}
In this section, we will first describe the benchmark dataset we created and then explain the proposed domain generation and evaluation pipeline as illustrated in Fig. \ref{fig:pipeline}.

\subsection{Dataset Generation}

\subsubsection{PDDL dataset}

We introduce a multi-domain PDDL dataset for UAV applications that consists of 30 domains, each representing a distinct UAV-related planning scenario. 
To reduce the risk of evaluating LLMs on data seen during training, we curate the domains from diverse sources, including academic literature on UAV planning, practical application studies, and modified versions of domains from the International Planning Competition (IPC)~\cite{ipc}.  
Unlike prior datasets that often rely on procedural, step-by-step instructional text~\cite{zhang-etal-2024-proc2pddl}, our domains are designed to capture the underlying structure of UAV environments, ensuring semantic coherence beyond surface-level instructions. 
In addition, several domains incorporate numeric fluents, extending beyond the classical STRIPS to model realistic UAV planning problems.  
Each domain is paired with 9 problem instances: 3 simple, 3 medium, and 3 hard. Simple problems can be solved with 1–2 actions, medium problems involve nearly the full action set, and hard problems require the complete set of actions while introducing additional complexity, such as multi-robot coordination.

\subsubsection{PDDL Domain Complexity Metric}
To systematically quantify the complexity of a PDDL domain, we introduce a composite metric that captures both structural and semantic characteristics of the domain. 
Specifically, the metric incorporates the following nine components:1) Number of actions, object types, predicates, and functions; 2) Average number of preconditions and effects per action; 3) Interdependency score: the average number of actions referencing a given predicate or function; 4) Action coupling: the number of conditions in an action's effects that appear as preconditions in other actions.
The final complexity score is computed as a weighted sum of each component.

Across our dataset, \textbf{domain complexity scores} range from 2.79 to 15.23, capturing a broad spectrum of difficulty—from simple navigation tasks to multi-agent coordination and numeric reasoning challenges. 
This diversity enables robust evaluation of LLMs across varying levels of domain complexity.
For reference, the well-known BLOCKSWORLD~\cite{Pellier02012018} domain (complexity: 5.23) ranks 17th out of 30, from most to least complex.
Based on this, we divide the dataset into 14 simple (\(\leq\) 5.23) and 16 complex (\(>\) 5.23) domains-a split used throughout our evaluation.

\subsection{Generate PDDL domain with LLMs}

We adopt an action-by-action strategy to generate each action within a domain. 
For action \(\hat{a_i}\), given the input tuple \( (N_d, N_{ai}, \text{Extern} ) \), we employ a coarse-to-fine approach to retrieve the most relevant example from our dataset for in-context learning. We then construct a prompt using a chain-of-thought method and feed it into LLM.
The resulting PDDL action is validated by a syntax checker, which provides feedback if syntax errors are detected. 
This process is repeated by re-prompting LLMs with feedback until either a syntactically correct output is generated or a maximum iteration number is reached.

\begin{tcolorbox}[colback=black!5!white, colframe=black!75!white, title=Example 1: Action Description Abstraction, 
fontupper=\scriptsize,
  left=2mm,   
  right=2mm,  
  top=1mm,    
  bottom=1mm  
  ]
\textbf{Original Action Description}
\vspace{-0.2em}
\begin{verbatim}
This action enables a uav to move from its current
position to a new position if two positions are
connected. This costs one unit of energy. For example,
the uav moves from position_1 to position_2 if there
is a connection between them. This costs one unit of
energy.
\end{verbatim}
\vspace{0.1em}
\textbf{After Abstraction}
\vspace{-0.2em}
\begin{verbatim}
This action enables a uav to move from [position1] to
[position2] if [position1] and [position2] are
[state1]. This action will increase [value1] by
[value2].
\end{verbatim}
\end{tcolorbox}

\subsubsection{Retrieval for Most Relevant In-context Example}
In-context learning is critical for enabling LLMs to perform structured generation tasks such as PDDL domain modeling, 
especially in zero-shot or low-resource settings.
Similarly to RAG, we utilize our dataset as an external source of domain-specific knowledge. 
We extract all actions from the dataset and convert each into an abstract representation using LLM to focus on semantic structures rather than lexical similarity, mitigating lexical noise.
Specifically, LLM replaces concrete nouns, adjectives, and numeric values with abstract placeholders such as \([object]\), \([state]\), and \([value]\),
while preserving subject and verb tokens to emphasize action semantics. 
LLM maintains a list of commonly used verbs.
If a verb in the description is semantically similar to one in the list, it is replaced.
Otherwise, it is added to the list and retained in the abstraction. 
We hypothesize that actions with similar semantic structures often correspond to similar PDDL formulations. 
The example below shows an action's original and abstracted forms, retaining key elements like `uav', `move', and `increase', while replacing other specifics with placeholders.
We encode all abstracted action representations into embeddings using the \texttt{all-mpnet-base-v2} model~\cite{all-mpnet-base-v2}, and store them in a searchable embedding index.

The retrieval process consists of two stages. 
In the coarse stage, we abstract the target action description \(N_{ai}\) and embed it as a query (Eg. 1); cosine similarities are computed across the stored index, and the top-$K$ most relevant examples are retrieved.
In the fine stage, these top-$K$ candidates are fed into LLMs to re-order and select the most semantically relevant example for in-context learning.

\subsubsection{Prompt with Chain of Thought}
To improve the reasoning capability of LLMs in PDDL domain generation, we employ a CoT prompting that directs the model to explicitly decompose the problem. 
The CoT process is embedded within context examples across three progressive reasoning steps.
First, we ask \texttt{What are the objects?}, prompting LLM to infer which objects are involved in the action based on the given context. 
This helps the model determine the correct action parameters.
Second, we ask \texttt{For each object, what are the preconditions? State them using predicates or functions.}
This encourages the model to reason about the conditions for action execution from the perspective of each object. 
Precondition can be either newly created or retrieved from an existing fluent list, 
and is paired with a natural language explanation to reinforce understanding.
Third, we ask \texttt{For each object, what are the effects? State them using predicates or functions.} which mirrors the second question but focuses on the state changes resulting from the action. 
This reinforces object-centered reasoning and ensures the model considers the outcome comprehensively.
Finally, we prompt \texttt{Write the action in the following format} to elicit a well-structured PDDL action.

\subsubsection{Iterative Generation of Actions}
Because the set of fluents is unknown in advance, we maintain a dynamic list of fluents throughout the generation process.
After generating each action, LLMs identify newly generated fluents, which are then added to the list and reused in subsequent actions.
This strategy mitigates LLM context length limitations and enables incremental domain construction, reducing redundant or inconsistent fluents.

Providing feedback to LLMs is key for iterative refinement.
While prior works rely on human\cite{guan2023leveraging} or environment-level feedback\cite{mahdavi2024leveraging}, these methods face scalability and accessibility challenges. 
Existing PDDL syntax checkers/parser also lack support for domains involving numeric.
Here, we extend a simple STRIPS-style syntax validator in \cite{guan2023leveraging} to support numeric checking and producing error feedback that can be directly used in the prompt, enabling a self-improvement loop.
For instance (Eg. 2), consider the erroneous precondition expression in the summation operator, where we can't add a predicate to a numerical function.

\begin{tcolorbox}[colback=black!5!white, colframe=black!75!white, title=Example 2: Syntax Checker, fontupper=\scriptsize,
  left=2mm,   
  right=2mm,  
  top=1mm,    
  bottom=1mm  
  ]
\textbf{Incorrect numerical input in precondition:}
\vspace{-0.2em}
\begin{verbatim}
(uav-number ?r - region); A function returning the 
                        ; number of UAV in region
(is-uav ?u - uav); A predicate indicating whether
                 ; ?u is a UAV
(+ (uav-number ?r) (is-uav ?u)) ; erroneous summation
\end{verbatim}
\vspace{0.1em}
\textbf{Syntax checker feedback:}
\vspace{-0.2em}
\begin{verbatim}
Head `at` in `Preconditions` is a predicate but should 
be a function. Please revise to fix this error. 
Note that you can always create new functions, but you
should also reuse existing functions whenever possible.
\end{verbatim}
\end{tcolorbox}

\section{Evaluation}
\label{set:Evaluation}
This section presents a series of evaluations conducted on our dataset to assess our prompt design. We aim to validate the following hypotheses:
\begin{itemize}
    \item \textbf{H1:} The proposed prompt designs will mitigate syntax errors, leading to a syntactically correct domain.
    \item \textbf{H2:} The proposed prompt designs will increase the semantic correctness of the generated domains.
\end{itemize}
Semantic correctness is assessed using three criteria: \textbf{1) Executability}, defined as the problem-solving success rate; \textbf{2) Feasibility}, which measures whether plans generated in a domain can be executed on the GT domain; and \textbf{3) Interpretability}, which measures whether the GT plan can be executed within the generated domain.

We evaluate two LLMs: GPT-4.1-mini-2025-04-14
(GPT) and DeepSeek-Chat
, using four existing prompting variations as benchmark methods. The compared methods and their abbreviations are: \textbf{1) Format}~\cite{guan2023leveraging}: Prompted with two regular examples for format referencing.
\textbf{2) Format+CoT (FCoT)}: Prompted with two regular examples for format referencing, both including chain-of-thought reasoning.
\textbf{3) Format+Semantic (FS)}: Prompted with one regular example for format referencing and one semantically relevant example for content referencing.
\textbf{4) Retrieval+CoT (Ours)}: Our framework retrieves one regular example for format referencing and one semantically relevant example for content referencing, both including chain-of-thought reasoning.

\vspace{-0.1cm}
\begin{table}[!h]
\centering
\renewcommand\arraystretch{1.1}
\caption{Domain categories in {\sc Spar}.}
\vspace{-0.2cm}
\begin{tabular}{p{2.2cm} p{5cm}}
\hline
\textbf{Category} & \textbf{Domains} \\
\hline
\textbf{Navigation} & 2DNavigation, HouseNavigation, \quad MultiCamLand, SimpleNavigation \\
\textbf{Transportation} & AirPassenger, CityCargo, Delivery, DeliveryCapacity, DeliveryDurative, DeliveryPriority, MedsDelivery, MedsDeliveryEnergy, MealDelivery\\
\textbf{Manipulation} & BlockWorld, CeilingPaint, GraspDelivery, GraspFruit, Watering \\
\textbf{Monitoring} & AirPollution,  ForestFireCollaboration,  MaintenanceComponent, MovableObs\\
\textbf{Exploration} & GridExplore, MarsExploration \\
\textbf{Aerial Operations} & Gate, Landing, Perching, Tracking, WindDisturbance  \\
\textbf{Adaptation} & TerrainAdapt, SpeedAdapt \\
\hline
\end{tabular}
\label{tab:env_groups}
\vspace{-0.2cm}
\end{table}

Given an input tuple \( (N_d, N_{A}, \text{Extern}) \) for domain \( \hat{D} \) in our dataset, we apply these four different methods to generate the corresponding PDDL domain \( \hat{D} \), respectively.
In total, we evaluate 30 domains, consisting of 14 simple (S.) domains and 16 complex (C.) domains as determined by \textbf{domain complexity scores}, and we summarize these domain categories in Tab.~\ref{tab:env_groups}.
The domains are grouped into seven categories: Navigation (autonomous flight), Transportation (logistics), Manipulation (object interactions), Monitoring (assessment), Exploration (coverage), Aerial Operations, and Adaptation. Together, they span tasks from low-level control to high-level decision-making.
Note that for \textbf{FS} and \textbf{Ours}, which treat the dataset as an external source and retrieve examples from it, we explicitly exclude all information from domain \( \hat{D} \) itself to prevent bias.
As shown in Fig. \ref{fig:pipeline}, after generating the domain, we evaluate it based on the number of syntax corrections, error types, problem-solving capability, and its overall feasibility and interpretability.

\vspace{-0.1cm}
\begin{table}[h]
\centering
\small
\renewcommand\arraystretch{1.1}
\setlength{\tabcolsep}{3.5pt}
\caption{Number of each error type across all domains.}
\vspace{-0.2cm}
\begin{tabular}{lcccc}
\multicolumn{5}{c}{\textbf{GPT-4.1-Mini}}\\
\toprule
\textbf{Error Type} 
& \begin{tabular}{@{}c@{}}\textbf{Format}\\(S./C./T.)\end{tabular}
& \begin{tabular}{@{}c@{}}\textbf{FCoT}\\(S./C./T.)\end{tabular}
& \begin{tabular}{@{}c@{}}\textbf{FS}\\(S./C./T.)\end{tabular}
& \begin{tabular}{@{}c@{}}\textbf{Ours}\\(S./C./T.)\end{tabular} \\
\midrule
object type      & 0/14/14 & 0/0/0    & 0/3/3    & 0/0/0 \\
predicate names  & 2/2/4   & 1/7/8    & 2/2/4    & 2/4/6 \\
predicate format & 1/5/6   & 1/5/6    & 0/1/1    & 0/1/1 \\
predicate usage  & 5/8/13  & 7/7/14   & 0/2/2    & 0/1/1 \\
function names   & 0/3/3   & 0/0/0    & 0/1/1    & 0/1/1 \\
function format  & 0/0/0   & 0/0/0    & 0/0/0    & 0/0/0 \\
function usage   & 0/0/0   & 0/3/3    & 1/3/4    & 0/13/13 \\
numeric usage    & 0/1/1   & 0/1/1    & 0/4/4    & 0/1/1 \\
Total error count & 8/33/41  & 9/23/32 & 3/16/19  & 2/21/23 \\
\bottomrule
\multicolumn{5}{c}{\textbf{DeepSeek-Chat}}\\
\toprule
\textbf{Error Type} 
& \begin{tabular}{@{}c@{}}\textbf{Format}\\(S./C./T.)\end{tabular}
& \begin{tabular}{@{}c@{}}\textbf{FCoT}\\(S./C./T.)\end{tabular}
& \begin{tabular}{@{}c@{}}\textbf{FS}\\(S./C./T.)\end{tabular}
& \begin{tabular}{@{}c@{}}\textbf{Ours}\\(S./C./T.)\end{tabular} \\
\midrule
object type      & 0/4/4   & 0/1/1   & 0/5/5   & 0/2/2 \\
predicate names  & 1/3/4   & 1/4/5   & 0/0/0   & 0/0/0 \\
predicate format & 6/11/17 & 1/5/6   & 2/1/3   & 0/1/1 \\
predicate usage  & 5/8/13  & 2/16/18 & 0/28/28 & 0/4/4 \\
function names   & 0/1/1   & 0/1/1   & 0/0/0   & 0/0/0 \\
function format  & 0/0/0   & 0/0/0   & 0/0/0   & 0/0/0 \\
function usage   & 1/12/13 & 1/2/3   & 0/2/2   & 0/1/1 \\
numeric usage    & 0/10/10 & 2/0/2   & 0/10/10 & 0/7/7 \\
Total error count & 13/49/62 & 7/29/36 & 2/46/48 & 0/15/15 \\
\bottomrule
\end{tabular}
\label{tab:error_count}
\vspace{-1em}
\end{table}

\begin{figure*}[!t]
      \centering
      \vspace{0.5em}
      \includegraphics[width=2\columnwidth]{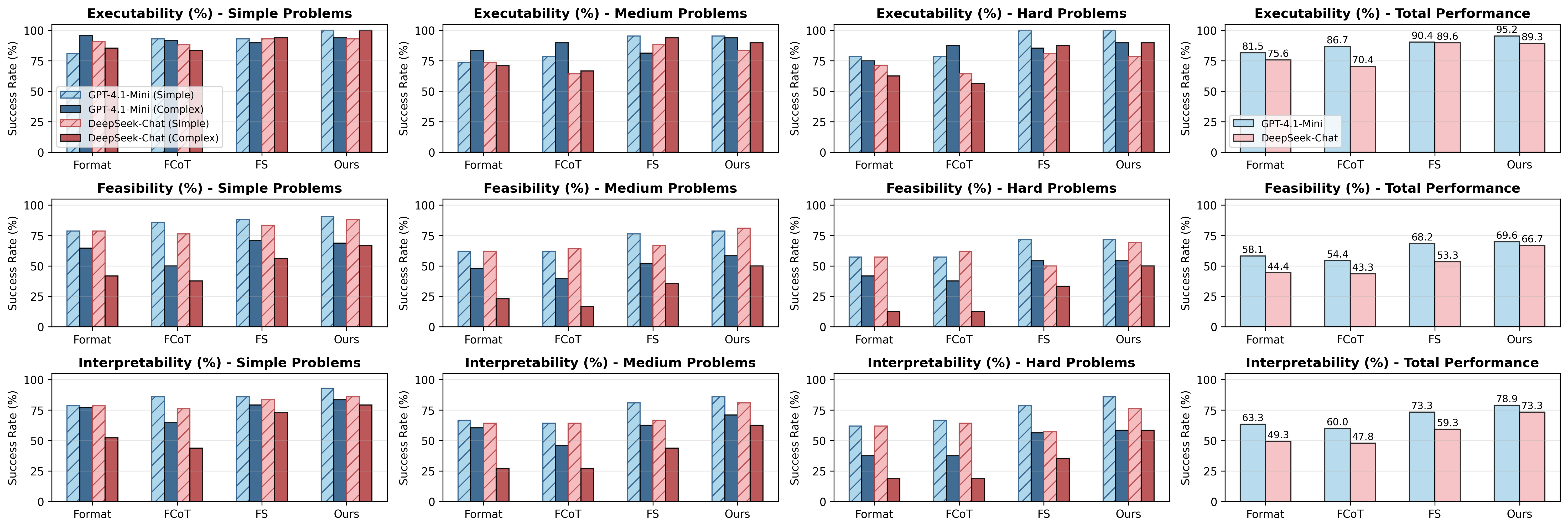}
      \vspace{-0.5em}
      \caption{Executability (\%), Feasibility (\%), and Interpretability (\%) scores for each method across 30 generated domains, and evaluated on a total of 270 problems (90 Simple, 90 Medium, and 90 Hard).}
     \label{img: performance}
      \vspace{-2em}
\end{figure*}

\subsection{Evaluation of Syntax Corrections and Error Types}

We inspect and summarize the following syntax error types and their correction number:  
\textbf{1) Object type}: The type in the action parameter does not match any type in the Extern input.  
\textbf{2) Predicate and function names}: The name of fluents conflicts with object type, existing predicate or function name, or PDDL keywords.  
\textbf{3) Predicate and function format}: Incorrect PDDL format inside predicates and functions.  
\textbf{4) Predicate and function usage}: Incorrect usage of predicates and functions with respect to their PDDL definitions.  
\textbf{5) Numeric usage}: Incorrect PDDL format and usage of numeric operators.

During generation, a syntax checker is employed to identify errors and provide feedback for LLMs to correct them. 
If the number of correction attempts exceeds the maximum allowed, we manually fix the remaining errors. 
The results are presented in Table~\ref{tab:error_count}.

Overall, GPT-4.1-Mini produces fewer total errors across all four methods compared to DeepSeek-Chat.
The dominant error types for GPT are object type, predicate usage, and function usage, whereas DeepSeek-Chat’s errors are primarily concentrated in predicate format, predicate usage, function usage, and numeric usage. 
This suggests that both models struggle with generating fluent expressions that are aligned with their definitions. 

Among benchmark methods, \textbf{Ours} consistently yields the relatively low number of errors across both LLMs, especially for predicate usage errors. 
Comparing \textbf{Ours} with \textbf{FCoT} and with \textbf{FS}, we observe that combining structured CoT reasoning with relevant in-context examples improves generation quality in terms of syntax, regardless of the model’s intrinsic ability.
In contrast, \textbf{Format} results in the highest error count for both LLMs, with a particularly high error count from DeepSeek-Chat. 
While \textbf{FS} achieves the lowest error count with GPT, it surprisingly exhibits a high number of predicate usage errors with DeepSeek-Chat. This instability may arise because the retrieved examples occasionally lack syntactic similarity to the target actions, leading DeepSeek-Chat to incorrectly transfer structural patterns from the examples to the current generation task.
\textbf{FCoT} also results in relatively high error counts. 
However, it notably reduces object type errors, suggesting that the reasoning steps in CoT prompts help the model recall type constraints more accurately. 

It is noteworthy that function-related syntax errors remain consistently absent across all methods and models, which can be attributed to the smaller and more uniform nature of function sets compared to predicates, making their structures easier for LLMs to reproduce reliably. 
In contrast, the majority of errors arise in complex domains, supporting the expectation that greater domain complexity increases the likelihood of syntactic mistakes.

\subsection{Evaluation of Semantic Correctness}

\subsubsection{Executability}
From the previous steps, the generated domains are ensured to be syntax-error free.
We then evaluate the problem-solving rate of all four methods across our dataset.
We first use LLMs to generate 9 problem files (3 simple, 3 medium, and 3 hard problems) for each generated domain.  
Each problem shares the same initial states and goals with its corresponding GT problem in our dataset.
We then feed both the generated domain and its generated problem files into a classical planner, the ENHSP-20 solver~\cite{Scala_Saetti_Serina_Gerevini_2020}, and report the results in the first row of Fig.~\ref {img: performance}. 
Overall, the GPT model outperforms DeepSeek-Chat in solving rate. 
The \textbf{FCoT} method improves performance by approximately 16\% when using GPT. 
In contrast, the \textbf{FS} shows limited variation across the two models. 
The solving rate generally decreases as the problem level increases and the domain complexity increases.
Among the four methods, \textbf{Ours} achieves the highest solving rate of 95.19\%, surpassing the \textbf{Format} method by about 14\%. 
The next best performers are \textbf{FS} and \textbf{FCoT}, suggesting that both chain-of-thought prompting and example retrieval contribute positively to solving performance, while their combination yields greater improvement.
However, successfully solving a problem merely indicates that a valid sequence of actions can be found from the initial state to the goal, i.e., the LLM-generated logic can be executable by a planner.
This does not imply that the solution is semantically meaningful or feasible for execution by a robot in the ground-truth domain.
Therefore, further evaluation of semantic correctness is essential.

\subsubsection{Feasibility and Interpretability}

To further assess the semantical correctness of the generated domains and verify whether the solved plans are feasible, we adopt a bidirectional domain validation strategy, inspired by~\cite{mahdavi2024leveraging, oswald2024largelanguagemodelsplanning}.
We treat the GT Domain as the optimal formulation of the environment,
and the GT Plan was solved from the GT Domain and GT Problem as a valid and feasible action sequence from the initial state to the final goal.
In the forward validation, we assess the feasibility of the generated domain by executing the plan (Gen Plan) produced in the generated domain on the GT Domain using VAL~\cite{VAL}.
If the Gen Plan can be executed on the GT Domain, it implies that the generated domain produces a feasible plan. 
The higher the number of feasible Gen Plans, the more reliable the generated domain is in terms of feasibility.

In inverse validation, we assess domain interpretability by executing the GT Plan within the generated domains (see Fig.~\ref{fig:pipeline}). 
Successful execution of the GT Plan indicates that the generated domain can reproduce the same action behaviors as the GT Domain, thereby demonstrating its interpretability. 
A higher number of executable GT Plans reflects greater interpretability of the generated domain.
Feasibility and interpretability results are presented in the second and third rows of Fig.~\ref{img: performance}. Overall, interpretability scores are consistently higher than feasibility scores across all generated domains. The highest feasibility and interpretability are achieved by the \textbf{Ours} method using GPT.
Notably, the \textbf{FCoT} method yields slightly lower feasibility and interpretability scores compared to the \textbf{Format}, indicating that CoT alone does not improve domain feasibility or interpretability. However, incorporating relevant examples (\textbf{FS} and \textbf{Ours}) significantly boosts performance in both metrics.

\section{Simulation Assets of {\sc Spar}}

\begin{figure}[!ht]
      \centering
      
      \includegraphics[width=1\columnwidth]{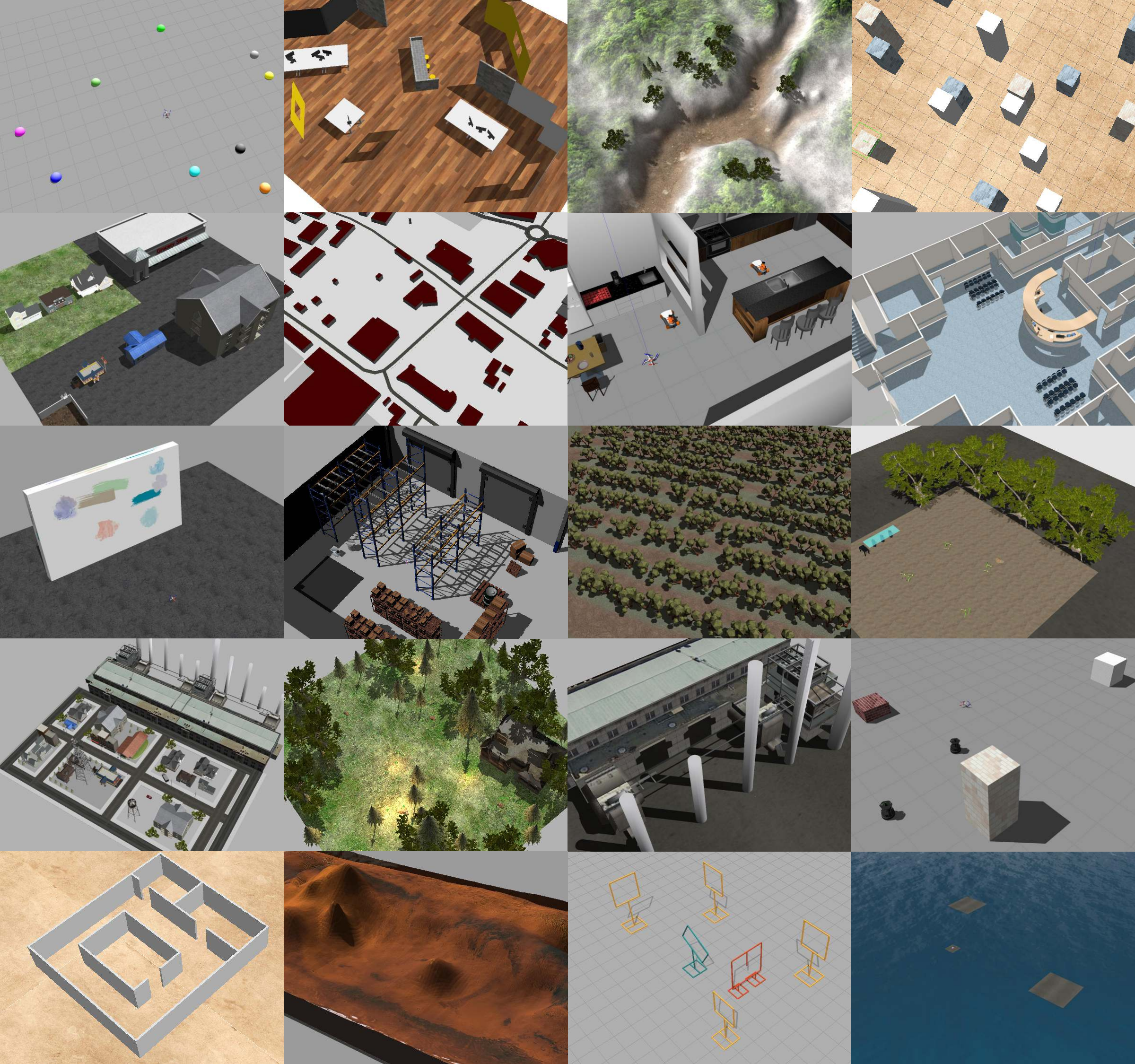}
      \vspace{-2em}
      \caption{Gazebo simulation environments across different domains (ordered as in Tab.~\ref{tab:env_groups}). Row 1: Navigation (2D, House, MultiCamLand, Simple). Row 2: Transportation. Row 3: Manipulation. Row 4: Monitoring. Row 5: Exploration (left two) and Aerial Operations (right two). The Adaptation domain combines variations of these environments, while WindDisturbance is modeled as an additional wind field independent of the environment.}
     \label{img: gazebo}
     \vspace{-1em}
\end{figure}

To test the domains generated by our framework in simulation, we customized a diverse set of 20 Gazebo environments spanning all domain categories described in Sec.~\ref{set:Evaluation}. 
This collection provides comprehensive coverage of tasks representative of real-world aerial robotic scenarios within a fully autonomous pipeline. Fig.~\ref{img: gazebo} shows the complete set of environments. 
{\sc Spar} is designed to be long-term and easily extensible: new environments and tasks can be incorporated as additional use cases arise. In this paper, we report on the current set of problems, with future releases to expand the repository accordingly.  
In the following, we illustrate the results using two complex generated domains as case studies.

\subsection{Case Study: UAV Watering tasks}

\begin{figure}[!t]
      \centering
      \vspace{0.5em}
      \includegraphics[width=1\columnwidth]{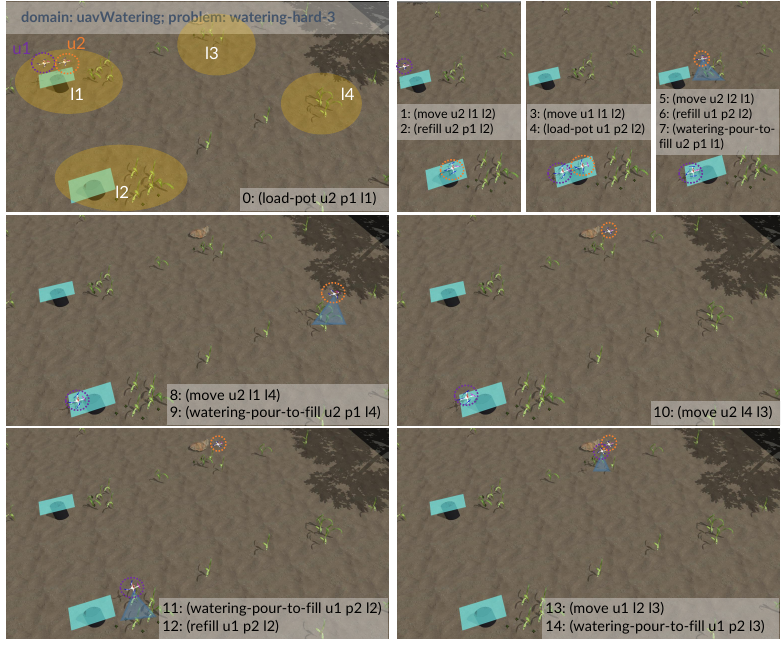}
      \vspace{-2em}
      \caption{Simulation of two UAVs serving four regions in the Watering domain. Example PDDL plan for watering-hard-3. UAVs u1 and u2 coordinate to load pots, move between locations, refill at water sources (l1 and l2), and perform sequential watering actions. The action \textit{watering-pour-to-fill} represents pouring water to meet a location’s demand.}
     \label{img: watering}
      \vspace{-1em}
\end{figure}

This domain models a UAV-based watering system where each UAV carries a single pot, refills at water sources, and pours water into target regions. The amount poured depends on the pot’s level and the location’s demand. The domain is extendable to multi-UAV settings.  
As shown in Fig.~\ref{img: watering}, two UAVs coordinate by loading pots, moving between locations, refilling at sources, and performing watering actions.

\subsection{Case Study: Multi-Robot Coordination}

\begin{figure}[!t]
      \centering
      \vspace{0.5em}
\includegraphics[width=1\columnwidth]{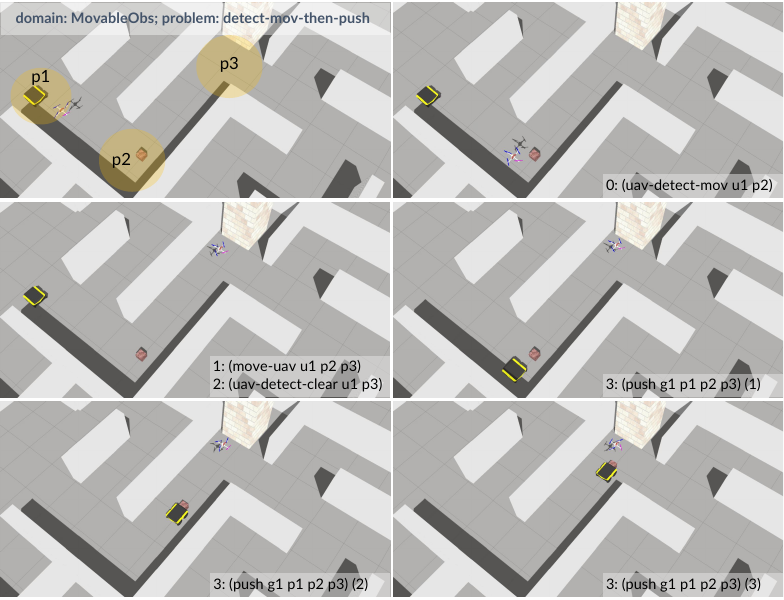}
      \vspace{-1em}
      \caption{Simulation of a detect-mov-then-push in the MovableObs domain. A UAV (u1) collaborates with a UGV (g1) to remove obstacles. The UAV detects a movable object at p2, verifies clearance at p3, and the UGV at p1 pushes the object from p2 to p3.}
     \label{img: moving}
      \vspace{-1em}
\end{figure}

This domain simulates a heterogeneous team of UAVs and unmanned ground vehicles (UGVs) in a discretized environment where positions may be free, blocked by fixed obstacles, or blocked by movable obstacles. UAVs can fly between positions to detect obstacles and determine their type, while UGVs can only traverse clear positions but are capable of pushing movable obstacles to adjacent locations to clear paths.  
We demonstrate a medium-level problem in Fig.~\ref{img: moving}, where a UAV detects a movable object, verifies clearance, and then enables the UGV to complete the push action for coordinated obstacle removal.

\section{Conclusion} 
\label{sec:conclusion}

In this work, we introduced a framework for automatic PDDL domain generation via LLMs, with UAV planning as a representative scenario. 
By developing a validated dataset of UAV domains and problems and designing a prompt framework tailored for domain generation, we demonstrated that LLMs can effectively produce domains that are both syntactically valid and semantically accurate. 
Our evaluations show that the generated domains not only satisfy structural requirements but also have strong feasibility and interpretability relative to ground-truth counterparts, enabling faster development, testing, and evaluation cycles.
The proposed dataset and pipeline also have significant potential for extending automatic domain generation to other robotic platforms, multi-agent planning, and real-world deployment, opening new opportunities at the intersection of language models and automated planning.

\bibliography{references}
\end{document}